\documentclass[twoside,11pt]{article}

\usepackage{blindtext}
\usepackage[utf8]{inputenc} % allow utf-8 input
\usepackage[T1]{fontenc}    % use 8-bit T1 fonts
\usepackage{url}            % simple URL typesetting
\usepackage{booktabs}       % professional-quality tables
\usepackage{amsfonts}       % blackboard math symbols
\usepackage{nicefrac}       % compact symbols for 1/2, etc.
\usepackage{microtype}      % microtypography
\usepackage{multirow}
\usepackage{xcolor}         % colors
\definecolor{codegreen}{rgb}{0,0.6,0}
\usepackage{graphicx}       % images
\usepackage{subfigure}		% sub figs
\usepackage{makecell}       % centering in table columns
\usepackage{amsmath}        % equation
\usepackage{xurl}           % url linebreak
\usepackage{lineno}  % line number
\usepackage{enumitem}

\usepackage{caption}
\usepackage{float}

\usepackage{listings}
\usepackage{simpleicons}

\usepackage{pifont}
\usepackage{xcolor}

\newcommand{\cmark}{\textcolor{green}{\ding{51}}} 
\newcommand{\xmark}{\textcolor{red}{\ding{55}}} 

\usepackage{algpseudocode}
\usepackage[algo2e,titlenotnumbered,boxed,ruled,vlined,linesnumbered]{algorithm2e}
\SetAlFnt{\small\sffamily}

\definecolor{mauve}{rgb}{0.58, 0.0, 0.83}

\usepackage{jmlr2e}

\usepackage{lastpage}
\jmlrheading{23}{2022}{1-\pageref{LastPage}}{1/21; Revised 5/22}{9/22}{21-0000}{Wenjie Du, Yiyuan Yang, Linglong Qian, Jun Wang, and Qingsong Wen}

\ShortHeadings{PyPOTS}{Du et al.}
\firstpageno{1}

\begin{document}

\title{PyPOTS: A Python Toolkit for Machine Learning on Partially-Observed Time Series}

\author{\name Wenjie Du \email wdu@time-series.ai \\
       \addr PyPOTS Research
       \AND
       \name Yiyuan Yang \email yiyuan.yang@cs.ox.ac.uk \\
       \addr PyPOTS Research \& University of Oxford
       \AND
       \name Linglong Qian \email linglong.qian@kcl.ac.uk \\
       \addr PyPOTS Research \& King’s College London
       \AND
       \name Jun Wang \email jwangfx@connect.ust.hk \\
       \addr PyPOTS Research \& Hong Kong University of Science and Technology
       \AND
       \name Qingsong Wen \email qingsongedu@gmail.com \\
       \addr Squirrel Ai Learning
}

\editor{My editor}

\maketitle

\begin{abstract}%   <- trailing '%' for backward compatibility of .sty file
\texttt{PyPOTS} is an open-source Python library dedicated to data mining and analysis on multivariate partially-observed time series with missing values. Particularly, it provides easy access to diverse algorithms categorized into five tasks: imputation, forecasting, anomaly detection, classification, and clustering. The included models represent a diverse set of methodological paradigms, offering a unified and well-documented interface suitable for both academic research and practical applications. With robustness and scalability in its design philosophy, best practices of software construction, for example, unit testing, continuous integration and continuous delivery, code coverage, maintainability evaluation, interactive tutorials, and parallelization, are carried out as principles during the development of \texttt{PyPOTS}. The toolbox is available on PyPI, Anaconda, and Docker. \texttt{PyPOTS} is open source and publicly available on GitHub~\url{https://github.com/WenjieDu/PyPOTS}.

\end{abstract}

\begin{keywords}
    time series, data science, machine learning, neural network, Python
\end{keywords}

\section{Introduction}

Missing data is a pervasive challenge in real-world time series due to sensor errors, communication failures, and other unexpected malfunctions. This issue leads to partially-observed time series (POTS), which can hinder advanced data analysis and modeling~\cite{wang2024deep}. Effective POTS algorithms are therefore highly valuable in many domains, including health monitoring~\cite{silva2012ICU}, network failure detection~\cite{du2021ILOS}, urban traffic forecasting~\cite{chen2021BTMF}, and gene expression analysis~\cite{bar2003Gene}.

To handle missing data in time series, several libraries offer imputation tools, as shown in Table~\ref{tab:main}, Python’s \texttt{ImputeBench}~\cite{khayati2020mind}, \texttt{Impyute}~\cite{law2021impyute}, \texttt{Autoimpute}~\cite{kearney2022autoimpute}, and \texttt{ImputeGAP}~\cite{nater2025imputegap}; and R’s \texttt{Mice}~\cite{vanbuuren2011mice} and \texttt{ImputeTS}~\cite{moritz2017imputeTS}.
While a two-step approach, imputing missing values followed by downstream modelling, can be effective, it may underperform in complex tasks. End-to-end methods that directly operate on POTS often yield better results. \texttt{PyPOTS} accommodates both strategies by supporting modular pipelines and integrated models~\cite{du2024tsi}.

Despite the importance of modeling partially-observed time series, there has been no dedicated library to support a wide range of data mining tasks on POTS, even in a community as vast as Python. \texttt{PyPOTS} was developed as a comprehensive Python toolbox to fill this gap. It provides end-to-end solutions for various tasks on POTS, moving beyond imputation-only methods to enable more reliable analysis of time series with missing data.

\begin{table*}[!t]
\caption{A comparison of available partially-observed time series (POTS) libraries. \cmark ~means implemented, (\cmark) indicates ongoing, and \xmark ~is unimplemented.}
\centering
\vspace{-10pt}
\resizebox{1\linewidth}{!}{
\renewcommand{\arraystretch}{1}{
\begin{tabular}{lcccccccc}
\toprule
\multirow{2}{*}{\textbf{Library}} & \multicolumn{5}{c}{\textbf{Modelling Task}} & \textbf{Number of} & \textbf{Number of} & \textbf{Execution} \\ \cmidrule(lr){2-6}
 & \textbf{Imputation} & \textbf{Forecasting} & \textbf{Detection} & \textbf{Classification} & \textbf{Clustering} & \textbf{Algorithms} & \textbf{Datasets} & \textbf{Environment}\\ \midrule
Mice & \cmark & \xmark & \xmark & \xmark & \xmark & 1 & 20 & R \\
ImputeTS & \cmark & \xmark & \xmark & \xmark & \xmark & 10 & 6 & R \\
Impyute & \cmark & \xmark & \xmark & \xmark & \xmark & 10 & 1 & Python \\
Autoimpute & \cmark & \xmark & \xmark & \xmark & \xmark & 8 & 1 & Python \\
ImputeBench & \cmark & \xmark & \xmark & \xmark & \xmark & 20 & 10 & Python \\
ImputeGAP & \cmark & (\cmark) & \xmark & \xmark & \xmark & 34 & 17 & Python \\ \midrule
\texttt{PyPOTS} & \cmark & \cmark & \cmark & \cmark & \cmark & \textbf{52} & \textbf{172} & Python \\ \bottomrule
\end{tabular}}}
\vspace{-15pt}
\label{tab:main}
\end{table*}

\texttt{PyPOTS} has the following evident advantages compared to existing packages: \textbf{1).} It contains 52 algorithms and 172 datasets that cover five modelling tasks on partially-observed time series; \textbf{2).} It provides a unified interface, detailed documentation, and interactive tutorials across all algorithms for ease of development and usage; \textbf{3).} It utilizes several automation services to measure, track, and ensure library quality, including cross-platform continuous integration with unit testing covering all algorithms, code coverage measurement, and maintainability evaluation; \textbf{4).} It employs optimization instruments whenever possible to enhance the scalability of the library. \texttt{PyPOTS} is capable of training models on large datasets but with limited computational resources, parallelly running a model across multiple GPU devices, and enabling all algorithms to train once and run anywhere.

\section{Project Concentration} \label{project}
\paragraph{Robustness Guarantee.}
GitHub actions are leveraged to automatically conduct unit testing with various versions of Python and different operating systems, including \textit{Linux (Ubuntu distribution)}, \textit{macOS}, and \textit{Windows}. There are CI (continuous integration) workflows that automatically run all tests daily and when a pull request is raised to ensure everything is good in the code base. When a new version with a batch of new features is released on GitHub, the CD (continuous delivery) workflows will automatically run the building pipelines to publish the new version on PyPI (Python Package Index), Anaconda, and Docker for delivering to users seamlessly.

\paragraph{Code Quality Assurance.}
\texttt{PyPOTS} project follows PEP 8 Python code style guide. Code coverage is published on and tracked by \href{https://coveralls.io}{\textit{Coveralls}}, a web-based code coverage service. And code maintainability is evaluated by \href{https://sonarcloud.io}{\textit{SonarQube Cloud}}, an automated tool for software quality assurance. \texttt{PyPOTS} currently has 85\% overall code coverage and 95\% maintainability (rated rank \textbf{A}). The code standards and these measurements are conducted to ensure the code quality and also to serve as safeguards for all pull requests from the open-source community.

\paragraph{Documentation and Tutorials.}
The comprehensive documentation is developed with \textit{Sphinx}, hosted on \href{https://readthedocs.org}{\textit{Read the Docs}}, and available on \href{https://docs.pypots.com}{https://docs.pypots.com}. It keeps the same docstring and rendering style as NumPy~\cite{harris2020numpy}. The documentation contains thorough installation instructions, quick-start examples, detailed API references, and an FAQ list. In addition to the documentation, \texttt{PyPOTS}' interactive tutorials in Jupyter Notebooks are released in the GitHub repository \href{https://github.com/WenjieDu/BrewPOTS}{https://github.com/WenjieDu/BrewPOTS}, and a simplified tutorial on \href{https://colab.research.google.com/drive/1HEFjylEy05-r47jRy0H9jiS_WhD0UWmQ}{\textit{Google Colab}} is provided for users to instantly run and explore.

\paragraph{Open-Source Community.}
As a practical machine learning library, \texttt{PyPOTS} receives extensive application within the scientific research community and has been recognized as a component of the \href{https://landscape.pytorch.org}{{\color[HTML]{EE4C2C}\simpleicon{pytorch}}\texttt{PyTorch Ecosystem}}. The code of \texttt{PyPOTS} is hosted on GitHub to be completely open source and to encourage collaborations from the community. We have community groups on instant message platforms (e.g., Slack and WeChat) for the community to promptly discuss and provide feedback. A dozen contributors are helping develop the framework, and others have contributed in the way of reporting bugs and requesting features.

\section{Design and Implementation} \label{design}
\paragraph{Base Frame.}
\texttt{PyPOTS} is built on top of common libraries like \texttt{NumPy}~\cite{harris2020numpy}, \texttt{Scikit-learn}~\cite{pedregosa2011sklearn}, \texttt{SciPy}~\cite{virtanen2020SciPy}, and \texttt{PyTorch}~\cite{paszke2019PyTorch} for modeling and computation, and uses \texttt{Pandas} and \texttt{H5py}~\cite{collette2023h5py} for data handling. Inspired by the API design of \texttt{Scikit-learn}~\cite{buitinck2013API}, it adopts a unified and task-oriented interface for imputation, forecasting, anomaly detection, classification, and clustering. \texttt{fit()} processes the training procedure and selects the best model checkpoint. \texttt{impute()}, \texttt{forecast()}, \texttt{detect()}, \texttt{classify()}, and \texttt{cluster()}, corresponding to each task, run inferences and return results. Unlike conventional libraries that rely on a generic \texttt{predict()} method, \texttt{PyPOTS} uses task-specific methods to clearly indicate model capabilities. Its modular design, leveraging inheritance and polymorphism, facilitates easy integration of new models.

\paragraph{Scalability Enhancement.}
To strengthen the scalability of \texttt{PyPOTS}, three key features are designed and implemented:
\underline{1). Data lazy-loading}: Industrial time-series datasets are often large and memory-intensive. To relieve this, \texttt{PyPOTS} provides a lazy-loading strategy in addition to full loading for small-sized datasets. Preprocessed data can be stored in HDF5 files, and only the necessary data batches are read into memory on demand during training and inference, significantly reducing memory usage;
\underline{2). Multi-device parallel acceleration}: Leveraging \texttt{PyTorch}, \texttt{PyPOTS} enables seamless GPU acceleration. For large datasets and compute-heavy models, users can scale across multiple GPUs simply by specifying device indices (e.g., \texttt{["cuda:0", "cuda:1"]}), making parallel training efficient and accessible;
\underline{3). Unified model serialization interface}: \texttt{PyPOTS} offers a consistent, model-agnostic, and device-agnostic interface for saving and loading models. This facilitates transferring trained models across devices and environments, supporting the "train once, run anywhere" paradigm and improving deployment flexibility.

\paragraph{Hyperparameter Optimization.}
Deep learning models rely heavily on hyperparameter choices, which significantly impact performance~\cite{feurer2019hyperparameter}. Given that most \texttt{PyPOTS} algorithms are neural networks, hyperparameter optimization is essential. This need is amplified by the diversity of time-series datasets across domains, where optimal settings for one dataset may perform poorly on another. To address this, \texttt{PyPOTS} integrates Microsoft \texttt{NNI}~\cite{MS2021NNI}, an AutoML toolkit for efficient hyperparameter tuning. Users only need to prepare two configuration files: one for defining the hyperparameter search space and the other for \texttt{NNI}'s tuning settings. Once launched, \texttt{PyPOTS} will communicate with NNI and provide a web interface to monitor the tuning process and view results, which can be sorted by performance metrics to identify the best configuration.

\paragraph{A Showcase.} 
The code snippet displays how to train a BRITS model to classify the PhysioNet-2012 dataset~\cite{Goldberger2000PhysioNet}. With a prepared dataset, users only need to write a few lines of code with \texttt{PyPOTS} to train a model and produce results on new data.
\begin{lstlisting}[
    language=Python, 
    label={code:showcasing}
]
# Our ecosystem kit will download and preprocess the dataset
from benchpots.datasets import preprocess_physionet2012
data = preprocess_physionet2012(subset='set-a',rate=0.1)
train_set = {"X": data["train_X"], "y": data["train_y"]}
val_set = {"X": data["val_X"], "y": data["val_y"]}
test_set, test_labels = {"X": data["test_X"]}, data["test_y"]

# Initialize and train the BRITS model
from pypots.classification import BRITS
from pypots.nn.functional import calc_binary_classification_metrics
model = BRITS(n_steps=data["n_steps"],
              n_features=data["n_features"],
              n_classes=data["n_classes"],
              rnn_hidden_size=256,
              epochs=5)
model.fit(train_set, val_set)

# Make predictions and evaluate
preds = model.classify(test_set)
metrics = calc_binary_classification_metrics(preds, test_labels)
\end{lstlisting}

\section{Conclusion} \label{conclusion}
This paper presents \texttt{PyPOTS}, a comprehensive Python toolkit for modeling partially-observed time series. It contains 52 algorithms and supports five tasks: imputation, classification, clustering, anomaly detection, and forecasting. Being committed to becoming a handy library, \texttt{PyPOTS} still has a long way to go in the future. At the time point of writing, most of the algorithms in \texttt{PyPOTS} are state-of-the-art methodologies based on neural networks, which can achieve outstanding results, but lacking explainability makes them not applicable in some sensitive fields, e.g., finance and marketing. We plan to include more models, especially ones with explainability, such as probabilistic methods and graph models. Besides, we will pay more attention to spatiotemporal data, which is also a common kind of time series, and implement models that work well with it.

\clearpage
\bibliography{references}

\end{document}